\documentclass[10pt,twocolumn,letterpaper]{article}

\usepackage[pagenumbers]{cvpr} 

\usepackage[dvipsnames]{xcolor}

\definecolor{cvprblue}{rgb}{0.21,0.49,0.74}
\usepackage[pagebackref,breaklinks,colorlinks,citecolor=cvprblue]{hyperref}
\usepackage{booktabs}       
\usepackage{amsfonts}       
\usepackage{nicefrac}       
\usepackage{pifont}
\usepackage{xcolor}         
\usepackage{multirow}
\usepackage{microtype}      
\usepackage{graphicx} 
\usepackage{amsmath}
\usepackage{amssymb}
\usepackage{enumitem}
\usepackage{wrapfig}
\usepackage{colortbl}
\usepackage{tcolorbox}
\usepackage{tikz}
\usetikzlibrary{calc}
\usepackage{url}

\newcommand{\myParagraph}[1]{\noindent \textbf{#1} ---}

\definecolor{TableGray1}{HTML}{9B9B9B}
\definecolor{TableGray2}{HTML}{C0C0C0}
\definecolor{TableGray3}{HTML}{EFEFEF}
\definecolor{TableGreen}{HTML}{BDDCAC}

\definecolor{ColBarSim}{HTML}{8A6729}
\definecolor{ColBarReal}{HTML}{367127}

\definecolor{Gray}{gray}{0.85}
\definecolor{GrayBorder}{gray}{0.65} 
\definecolor{green_cylinder}{rgb}{0.0,0.40,0.0}
\definecolor{blue_cylinder}{rgb}{0.02,0.05,0.75}
\definecolor{way_point}{rgb}{0.56,0.0,1.0}
\newcolumntype{C}{>{\columncolor{TableGray2}}c}
\newcolumntype{L}{>{\columncolor{TableGray2}}l}
\newcolumntype{L}{>{\columncolor{TableGray2}}l}
\newcolumntype{R}{>{\columncolor{TableGray2}}r}
\newcolumntype{S}{>{\columncolor{blue!20}}r}
\newcolumntype{N}{>{\columncolor{orange!20}}r}
\newcolumntype{R}{>{\columncolor{green!20}}r}

\newcommand{\xmark}{\textcolor{black!50}{\ding{55}}}
\newcommand{\cmark}{\ding{51}}

\newcommand\myparagraph[1]{\textbf{#1} ---}

\newcommand{\epbox}[1]{\tcbox[on line,colframe=black,boxsep=0pt,left=1pt,right=1pt,top=1pt,bottom=1pt,boxrule=0.4pt]{#1}}

\def\paperID{339} 
\def\confName{CVPR}
\def\confYear{2024}

\title{Learning to navigate efficiently and precisely in real environments}

\author{Guillaume Bono, Herv\'e Poirier, Leonid Antsfeld, Gianluca Monaci, Boris Chidlovskii, Christian Wolf\\ \\
Naver Labs Europe, Meylan, France\\
{\tt\small \{firstname\}.\{lastname\}@naverlabs.com}
}

\begin{document}

\maketitle

\begin{abstract}
\noindent
In the context of autonomous navigation of terrestrial robots, the creation of realistic models for agent dynamics and sensing is a widespread habit in the robotics literature and in commercial applications, where they are used for model based control and/or for localization and mapping. The more recent Embodied AI literature, on the other hand, focuses on modular or end-to-end agents trained in simulators like Habitat or AI-Thor, where the emphasis is put on photo-realistic rendering and scene diversity, but high-fidelity robot motion is assigned a less privileged role. The resulting sim2real gap significantly impacts transfer of the trained models to real robotic platforms.
In this work we explore end-to-end training of  agents in simulation in settings which minimize the sim2real gap both, in sensing and in actuation. Our agent directly predicts (discretized) velocity commands, which are maintained through closed-loop control in the real robot. The behavior of the real robot (including the underlying low-level controller) is identified and simulated in a modified Habitat simulator. Noise models for odometry and localization further contribute in lowering the sim2real gap. We evaluate on real navigation scenarios, explore different localization and point goal calculation methods and report significant gains in performance and robustness compared to prior work.
\end{abstract}

\begin{figure}[t] \centering
    \includegraphics[width=\linewidth]{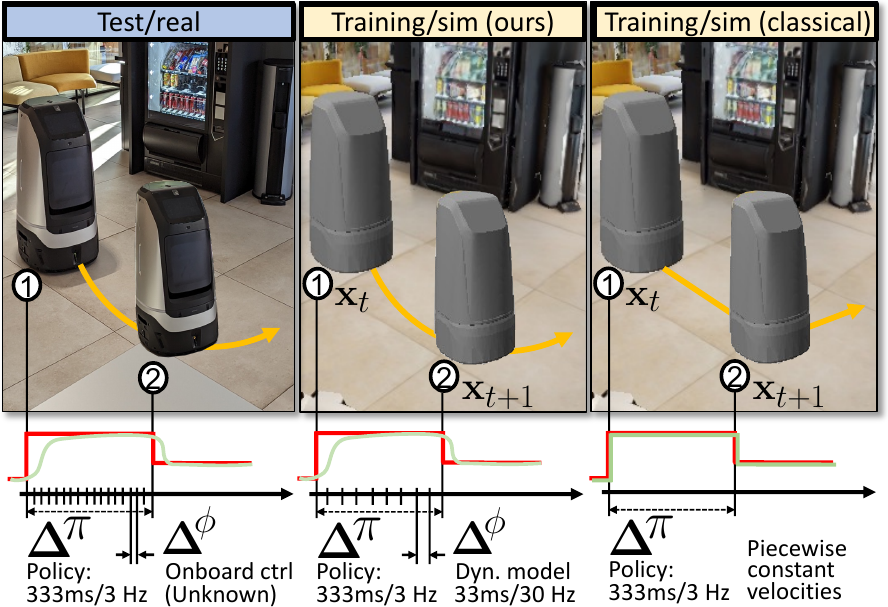}
    \caption{\label{fig:teaser}
    Efficient navigation with policies end-to-end trained in 3D photorealistic simulators requires closing the sim2real gap in sensing \emph{and} actuation.    
    Efficiency demands that the robot continues to move during decision taking (as opposed to stopping for each sensing operation), and this requires a realistic motion model in simulation allowing the agent to internally anticipate its future state. This requirement is exacerbated by the delay between sensing \ding{172} and actuation \ding{173} caused by the computational complexity of high-capacity deep networks (visual encoders, policy). To model realistic motion while training in simulation, we create a $2^{nd}$ order dynamical model running with higher frequency, which models the robot and its low-level closed-loop controller. We identify the model from real data and add it to the Habitat~\cite{Savva_2019_ICCV} Simulator.}    
\end{figure}

\section{Introduction}
\label{sec:introduction}

\noindent
Point goal navigation of terrestrial robots in indoor buildings has traditionally been addressed in the robotics community with mapping and planning \cite{thrun2005probabilistic,bresson2017simultaneous,lluvia_active_2021}, which led to solutions capable of operating on robots in real environments. The field of computer vision and embodied AI has addressed this problem through large-scale machine learning in simulated 3D environments from reward with RL \cite{DBLP:conf/iclr/MirowskiPVSBBDG17,DBLP:conf/iclr/JaderbergMCSLSK17} or with imitation learning \citep{DBLP:conf/nips/DingFAP19}. Learning from large-scale data allows the agent to pick up more complex regularities, to process more subtle cues and therefore (in principle) to be more robust, to exploit data regularities to infer hidden and occluded information, and generally to learn more powerful decision rules. In this context, the specific task of point goal navigation (navigation to coordinates) is now sometimes considered ``solved'' in the literature 
\cite{IsMappingNecessaryCVPR2022}, incorrectly, as we argue. While the machine learning and computer vision community turns its attention towards the exciting goal of integrating language models into navigation \cite{huang2022innermonologue,driess2023palme}, we think that further improvements are required to make trained agents perform reliably in real environments with sufficient speed.

Experiments and evaluations of trained models in real environments and the impact of the sim2real gap do exist in the Embodied AI literature \cite{chattopadhyay2021robustnav,SadekICRA2022,gervet2023navigating}, but they are rare and were performed in restricted environments. Known models are trained in 3D simulators like Habitat \cite{Savva_2019_ICCV} or AI-Thor \cite{ai2thor}, which are realistic in their perception aspects, but lack the accurate motion models of physical agents which the robotics community uses for articulated robots, for instance, where forward and inverse kinetics are often modeled and identified. This results in a large sim2real gap in motion characteristics, which is either ignored or addressed by discrete motion commands. The latter strategy teleports the agent in simulation for each discrete action and executes these actions on the physical robot by low-level control on positions instead of velocities, stopping the robot between agent decisions and leading to extremely inefficient navigation and overly long episodes \cite{SadekICRA2022}.

In this work we go beyond current models for terrestrial navigation by realistically modeling both actuation and sensing of real agents in simulation, and we successfully transfer trained models to a real robotic platform. On the actuation side, we take agent decisions as discretized linear and angular velocities and predict them asynchronously while the agent is in motion. The resulting delay between sensing and actuation (see Figure \ref{fig:teaser}) requires the agent to anticipate motion, which it learns in simulation from a realistic motion model. To this end, we create a second-order dynamical model, which approximates the physical properties of the real robot together with its closed-loop low-level controller. We identify this model from real robot rollouts and expert actions. The model is integrated into the Habitat simulator as an intermediate solution between the standard instantaneous motion model and a full-blown computationally-expensive physics simulation based on forces, masses, frictions etc.

On the sensing side, we combine visual and Lidar observations with different ways of communicating the navigation (point) goal to the agent, and different estimates of robot pose relative to a reference frame, combining two different types of localization: first, dead reckoning from wheel encoders, and secondly, external localization based on Monte Carlo methods from Lidar input. We perform extensive experiments and ablations on a real navigation platform and quantify the impact of the motion model, action spaces, sensing capabilities and the way we provide goal coordinates on navigation performance, showing that end-to-end trained agents can robustly navigate from visual input when their motion has been correctly modelled in simulation.

\section{Related Work}
\label{sec:relatedwork}

\noindent
\myparagraph{Visual navigation} navigation has been classically solved in robotics using mapping and planning~\citep{burgard1998interactive,macenski2020marathon,marder2010office}, which requires solutions for mapping and localization~\citep{bresson2017simultaneous, labbe19rtabmap,thrun2005probabilistic}, 
for planning~\citep{konolige2000gradient, sethian1996fast} and for low-level control \citep{fox1997dynamic,rosmann2015timed}. These methods depend on accurate sensor models, filtering, dynamical models and optimization. End-to-end trained models directly map input to actions and are typically trained with RL~\citep{DBLP:conf/iclr/JaderbergMCSLSK17,mirowski17learning} or imitation learning~\citep{DBLP:conf/nips/DingFAP19}. They learn representations such as flat recurrent states, occupancy maps~\citep{Chaplot2020Learning}, semantic maps~\citep{chaplot2020object, gervet2023navigating}, latent metric maps~\citep{DBLP:conf/pkdd/BeechingD0020,Henriques_2018_CVPR,DBLP:conf/iclr/ParisottoS18}, topological maps~\citep{BeechingECCV2020,Chaplot_2020_CVPR,shah2022viking}, self-attention~\citep{chen_think_2022,du2021vtnet,Fang_2019_CVPR,reed_generalist_2022}, implicit representations~\citep{Marza2022NERF} or maximizing navigability directly with a blind agent optimizing a collected representation~\cite{Mole2024}. 
Our proposed method is end-to-end trained and features recurrent memory, but benefits from an additional motion model in simulation. The perception modules (visual encoders) required in visual navigation have traditionally increased in capacity over time, starting with small-capacity CNNs with a couple of layers, moving to ResNet variants (eg. in \cite{wijmans2019dd}), then to ViTs (eg. in \cite{majumdar2023search,yadav2023ovrlv2}) and to binocular ViTs for the case of ImageGoal (eg. in \cite{CrocoNav2024}).

\noindent
\myparagraph{Sim2real transfer}
The sim2real gap can compromise strong performance achieved in simulation when agents are tested in the real-world~\cite{hofer2020sim2real,liu2023world}, as perception and control policies often do not generalize well to real robots due to inaccuracies in modelling, simplifications and biases. To close the gap, domain randomization methods~\cite{peng18sim,tan18real} treat the discrepancy between the domains as variability in the simulation parameters. 
Alternatively, domain adaptation methods learn an invariant mapping for matching distributions between the simulator and the robot environment. Examples include transfer of visuo-motor policies by adversarial learning~\cite{zang19adversarial}, adapting dynamics in RL~\cite{eysenbach21off}, lowering fidelity simulation~\cite{truong_rethinking_2022} and adapting object recognition to new domains~\cite{zhu19adapting}. Bi-directional adaptation is proposed in~\cite{truong_bi-directional_2021}; Recently, \citet{chattopadhyay2021robustnav} benchmarked the robustness of embodied agents to visual and dynamics corruptions.
\citet{kadian_sim2real_2020} investigated sim2real predictability of Habitat-Sim~\cite{Savva_2019_ICCV} for PointGoal navigation and proposed a new metric to quantify it, called Sim-vs-Real Correlation Coefﬁcient (SRCC). The PointGoal task on real robots is also evaluated in~\cite{SadekICRA2022}. To reach competitive performance without modelling any sim2real transfer, the agent is pre-trained on a wide variety of environments and then fine-tuned on a simulated version of the target environment. 

\noindent
\myParagraph{Hybrid methods}
Modular and hybrid approaches decompose planning into different parts, frequently in a hierarchical way. Typically, waypoints are proposed by a high-level (HL) planner, and then followed by a low-level (LL) planner, as in \cite{Chaplot2020Learning,Chaplot2020semantic,SadekICRA2022,bansal2020combining,BeechingGamerland2022, gervet2023navigating}. Hybrid methods combine classical planning with learned planning. Some of the modular approaches mentioned above can be considered to be hybrid, but there also exist hybrid approaches in the literature which combine different planners more tightly and in a less modular way~\cite{Sim2RealStrategy,dashora2022hybrid,neuralastar2021,gupta2017cognitive,beeching_learning_2020}. In some of these cases, training back-propagates through the planning stage. Some methods combine the advantages of classical and learned planning by switching between them~\cite{kastner2022all,SombitIROS2023}. These methods also aim to decrease the sim2real gap and are complementary to ours.

\noindent
\myparagraph{Motion models}
are usually kept very simple when training visual navigation policies. 
Several works use a discrete high-level action space in simulation~\cite{chaplot2020object, Chaplot2020Learning, gervet2023navigating} and rely on dedicated controllers to translate these discrete actions into robot controls when transferring to the real world. Alternative approaches predict continuous actions~\cite{choi2019deep, yokoyama2022learning} or waypoints~\cite{bansal2020combining, shah2023vint}, that again are subsequently mapped to robot controls when executing the policies in real. Somewhat closer to our approach, \citet{yokoyama2021success} adopt a discretized linear and angular velocity space, and introduce a new metric, SCT, that uses a simplified ``unicycle-cart'' dynamical model to estimate the ``optimal'' episode completion time computed by rapidly exploring random trees (RRT).
However the unicycle model is only used for evaluation and it is not deployed for training nor testing: as in previous works, the commands are mapped to robot controls using handcrafted heuristics. In contrast to previous approaches, we include the dynamical model of the robot in the training loop to learn complex dependencies between the navigation policy and the robot physical properties.

\section{End-to-end training with realistic sensing}
\label{sec:method}

\begin{figure}[t] \centering
    \includegraphics[width=\linewidth]{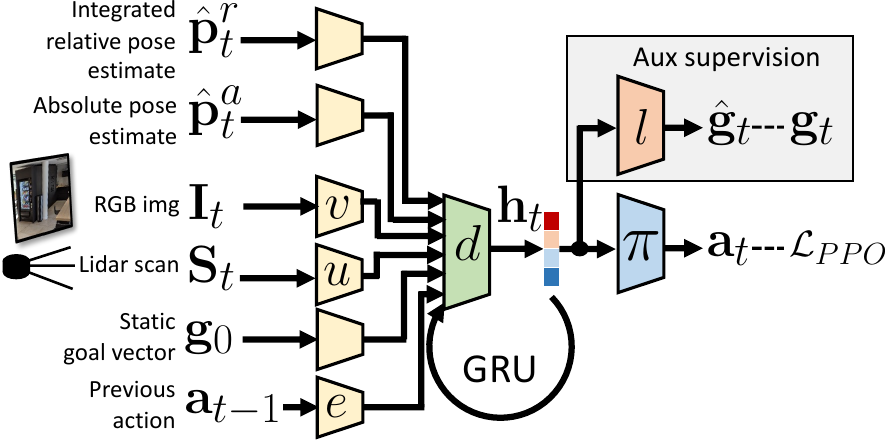}
    \caption{\label{fig:policy}\textbf{The agent} uses a recurrent policy with a static point goal $g_0$ as input, i.e. the goal is constant and given wrt. to the initial reference frame. During training, the estimation of the dynamic point goal $\hat{g}_t$ is supervised from privileged information. 
    }    
    \vspace*{-3mm}
\end{figure}

\noindent
We propose a method for training agents which (in principle) could address a variety of visual navigation problems, and without loss of generality, focusing on sim2real transfer and efficiency, we formalize the problem as a \textit{PointGoal} task: an agent receives sensor observations at each time step $t$ and must take actions $\mathbf{a}_t$ to reach a goal position given as polar coordinates.
The method is not restricted to any specific observations, in our experiments the agent receives RGB images $\mathbf{I}_t {\in}\mathbb{R}^{3\times H\times W}$ and a Lidar-like vector of ranges $\mathbf{S}_t {\in}\mathbb{R}^{K}$, where $K$ is the number of laser rays.

\begin{figure*}[t] \centering
    \includegraphics[width=\linewidth]{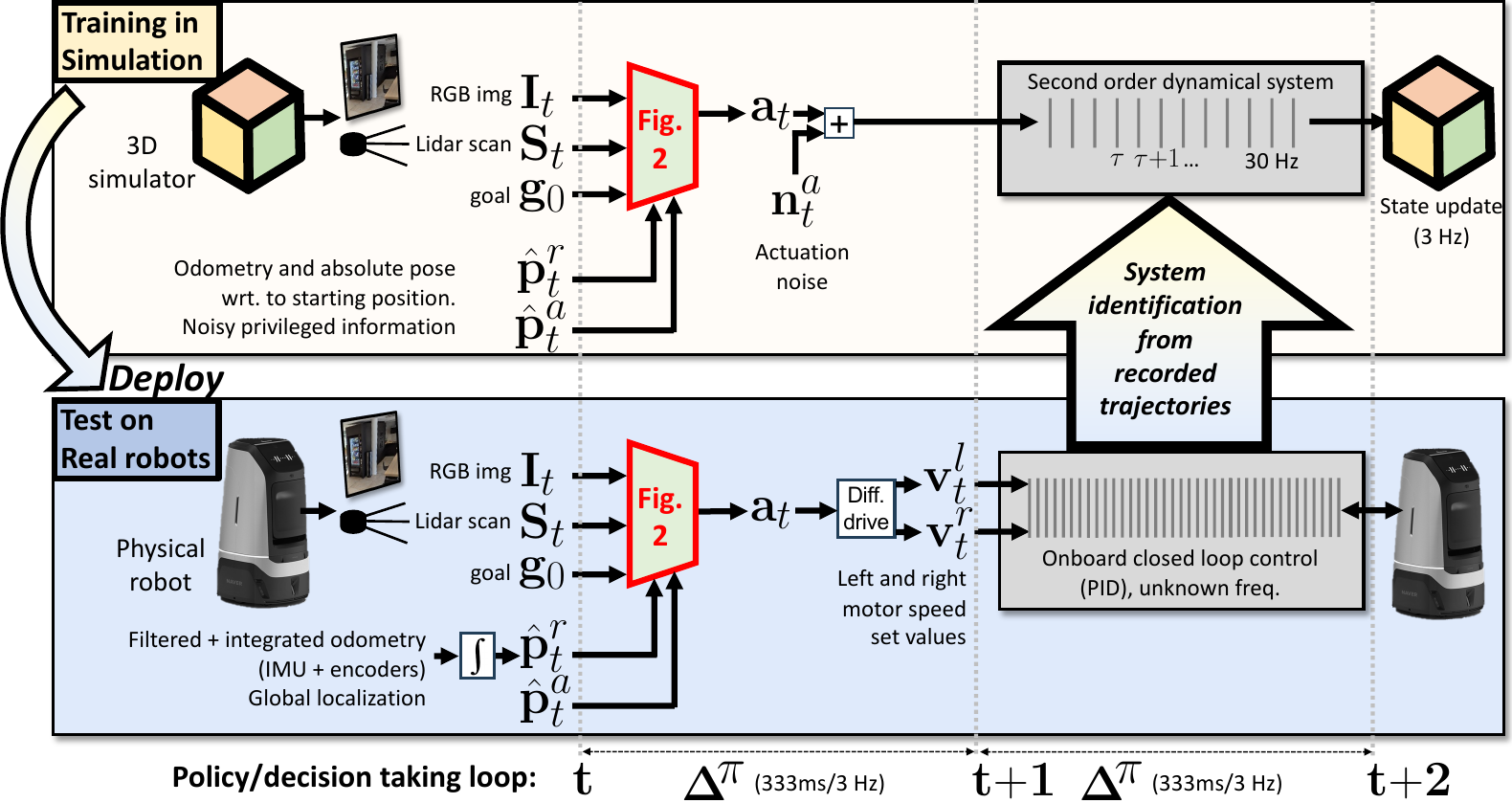}    \caption{\label{fig:schema}\textbf{Training visual navigation with realistic motion models:} we train an end-to-end agent in simulation (top) subject to two different simulation loops: a slower loop at 3Hz (indexed by $t$) renders visual observations and takes agent decisions, while a faster loop at 30 Hz (indexed by $\tau$) simulates physics. Physics is approximated with a $2^{nd}$ order model identified from real robot rollouts (bottom) and includes the robot physics as well as the behavior of the closed-loop control of the differential drive (neither onboard control algorithm nor control frequency need to be known). Operations in the intervals are pipelined, eg. sensing occurs at each time step, as does agent forward pass etc. The agent architecture is detailed in Figure~\ref{fig:policy}.} 
    \vspace*{-5mm}
\end{figure*}

The proposed agent is fully trained and therefore does not require to localize itself on a map for analytical path planning. However, even in this very permissive setting the application itself requires to communicate the goal to the agent in some form of coordinate system, which requires localization at least initially with respect to the goal (in \textit{ImageGoal} settings this would be replaced by the goal image). We can classically differentiate between three different forms:
\begin{description}[labelindent=2mm,leftmargin=4mm,labelsep=0.5em,itemsep=1mm,nolistsep]
    \item[Absolute PointGoal ] --- the goal vector $\mathbf{g}^a = [ \mathbf{g}^a_x, \mathbf{g}^a_y ]^T$ is given in a pre-defined agent-independent absolute reference frame. This requires access to a global localization system.
    \item[Dynamic PointGoal] --- the goal vector $\mathbf{g}_t$ is provided with respect to the \textit{current} agent pose and therefore updated at each time instant $t$. This requires an external localization system providing estimates which are then transformed into the egocentric reference frame.
    \item[Static PointGoal] --- the goal vector is provided with respect to the starting position at $t{=}0$ of the robot, and not updated with agent motion (=$\mathbf{g}_0$). This requires the agent to learn this update itself, and therefore some form of internal localization or integration of odometry information.    
\end{description}
We target the \textit{Static PointGoal} case, which does not require localization beyond the initial relative vector. However, we also argue that some form of localization and/or odometry information is useful, which will assist the dynamics information inferred from (visual) sensor readings. Furthermore, during training it will allow the agent to learn a useful latent representation in its hidden state and an associated dynamics in this latent space. We therefore provide the agent with two different localization components:
\begin{description}[labelindent=2mm,leftmargin=8mm]
    \item[$\hat{\mathbf{p}}^a_t$] --- an estimate of agent pose with respect to the initial agent position, asynchronously delivered and with low frequency, around 0.5-1 fps. At test, this signal is delivered by external localization, in our case from a Lidar-based AMCL module (see Section \ref{sec:experiments}).
    \item[$\hat{\mathbf{p}}^r_t$] --- an estimate coming from onboard odometry, which we integrate over time to provide an estimate situated in the same reference frame as $\hat{\mathbf{p}}^a_t$, i.e. with respect to the initial agent position.
\end{description}
The characteristics of the two localization signals are different: while $\hat{\mathbf{p}}^r_t$ corresponds to a \textit{dead reckoning} process and is subject to drift, its relative noise is lower. The signal $\hat{\mathbf{p}}^a_t$ is not subject to drift but suffers from larger errors in certain regions, due to registration errors or absence of 
structure. 


The agent policy is recurrent, based on a hidden memory vector $\mathbf{h}_t$ updated over time from input of all sensor readings $\mathbf{I}_t, \mathbf{S}_t, \mathbf{p}^r_t, \hat{\mathbf{p}}^a_t$, the goal $\mathbf{g}_0$ and the previous action $\mathbf{a}_{t-1}$,
\begin{align}
    \mathbf{h}_t = & d(\mathbf{h}_{t-1}, v(\mathbf{I}_t), u(\mathbf{S}_t), \mathbf{g}_0, \hat{\mathbf{p}}^r_t, \hat{\mathbf{p}}^a_t, e(\mathbf{a}_{t-1})),
    \label{eq:gru}
    \\
    \mathbf{a}_t = & \pi(\mathbf{h}_t), 
    \label{eq:policy} \\
    \hat{\mathbf{g}}_t = & l(\mathbf{h}_{t})
    \label{eq:supervision}
    , 
\end{align}
where $d$ is a GRU with hidden state $\mathbf{h}_t$, $v$ is a visual encoder of the image $\mathbf{I}_t$ (in our case, a ResNet-18); $u$ is a 1D-CNN encoding the scan $\mathbf{S}_t$; $e$ is a set of action embedding vectors; $\pi$ is a linear policy. Other inputs go through dedicated MLPs, not named to lighten notations.
To help the agent deal with the noisy relative and absolute pose estimates, we put an inductive bias for localization on $\mathbf{h}_t$ through an auxiliary head $l$ predicting the dynamic goal compass $\hat{\mathbf{g}}_t$, supervised during training from the simulated agent's ground truth pose, as shown in Figure~\ref{fig:policy}. In the following paragraphs we describe the two localization systems and the noise models we use to simulate them during training in simulation. Section~\ref{sec:dynamicalmodel} then introduces the dynamical model used during training.

\myparagraph{Integrated relative localization}
the pose estimate $\hat{\mathbf{p}}^r_t$ can be obtained with any commercially available odometry system. In our experiments we integrate readings from wheel encoders.
During simulation and training, $\hat{\mathbf{p}}^r_t$ is estimated by sampling planar noise $[\epsilon_x, \epsilon_\theta]^\intercal$ from a multivariate Gaussian distribution 
$$
\mathcal{N}(\begin{bmatrix}0.01 \\ 0\end{bmatrix}, \begin{bmatrix}10^{-4} & 10^{-4} \\ 10^{-4} & 10^{-3}\end{bmatrix}),
$$
which gets added to the step-wise motion of the agent, leading to the accumulation of drift over time.

\myparagraph{Absolute localization}
the pose estimate $\hat{\mathbf{p}}^a_t$ is obtained with the standard ROS 2 package for Adaptive Monte Carlo Localization (AMCL) \cite{thrun2005probabilistic}, a KLD-sampling approach requiring a pre-computed, static map based on particle filtering. 
During simulation and training, $\hat{\mathbf{p}}^r_t$ is estimated by sampling planar noise from another Gaussian $\mathcal{N}(\mathbf{0}, \mathbf{Diag}[0.03,0.03, 0.05])$, but this time it gets directly added to the ground truth pose of the robot, thus modeling imprecise localization (eg. due to Lidar registration failures) without drift accumulation.
In addition, we model the potential unreliability of such absolute pose estimates by only providing a new observation to the agent every few steps, at an irregular period governed by a uniform distribution $\mathcal{U}(8, 12)$. This potentially allows to plug-in low-frequency visual localization.

\section{A dynamical model of realistic robot motion}
\label{sec:dynamicalmodel}
\noindent
One of our main motivations is to be able to use a trained model to navigate \emph{efficiently}, i.e. fast and reliably. Stopping the robot between predicted actions is therefore not an option, and we train the model to predict linear and angular velocity commands in parallel to its physical motion, as illustrated in Figure \ref{fig:teaser}. This requires the agent to anticipate motion to predict an action $\mathbf{a}_t$ relative to the next (yet unobserved) state at $t$, exacerbated by the delay between sensing at time $t$ and actuation at $t{+}1$, during which computation through the high-capacity visual encoder and policy happens.
This motion depends on the physical characteristics of the robot, as shown in Figure \ref{fig:schema} (bottom):  the linear and angular velocity commands predicted by the neural policy $\pi$ are translated into ``set values'' for left and right motor speeds of the differential drive system with a deterministic function and used as targets by the closed loop control algorithm, typically a PID (\textit{Proportional Integral Derivative Controller}), attempting to reach and maintain the predicted speed between agent decisions. The behavior depends on the control algorithm but also on the physical properties of the robot, like its mass and resulting inertia, power and torque curves of the motors, friction etc. 
We decrease the actuation sim2real gap as as much as possible by integrating a model of realistic robot behavior into the Habitat~\cite{Savva_2019_ICCV} platform, shown in the top part of Figure~\ref{fig:schema}. We approximate the robot's motion behavior by modeling the \textit{combined} behavior of both the robot and the control algorithm implemented on the real platform as an asymmetric second-order dynamic system. The parameters of this low-level loop are estimated from recordings of the real robot platform with a system identification algorithm.

Classically, a sequential decision process can be modeled as a POMDP $\langle\mathcal{X},\mathcal{U},T,\mathcal{O},O,R\rangle$.
In pure 2D navigation settings with a fixed goal in a single static scene, the environment maintains a (hidden) state $\mathbf{x}_t\in\mathcal{X}$ consisting in the pose of the agent, i.e. position and orientation, $\mathbf{x}_t = [ x_t, y_t, \theta_t ]$. In most settings in the literature based on 3D photo-realistic simulators like Habitat \cite{Savva_2019_ICCV}, the state update in simulation ignores physical properties of the robot (mass/inertia, friction, accelerations,  etc.), as it is implemented through ``teleportation'' of the robot to a new pose. In that case, 
the environment transition decomposes 
into a discrete state update $x_{t+1} = T(x_t, u_t)$ and an observation $o_{t+1}=O(x_{t+1})$,
The discrete action space $\mathcal{U}_\text{motion} = \{${\small \texttt{FORWARD 25cm}, \texttt{TURN\_LEFT $10^{\circ}$}, \texttt{TURN\_RIGHT $10^{\circ}$} and \texttt{STOP}}$\}$ frequently chosen in the navigation literature results in a halting and jerky robot motion.

\begin{wrapfigure}{r}{3.1cm}
\includegraphics[width=3.1cm]{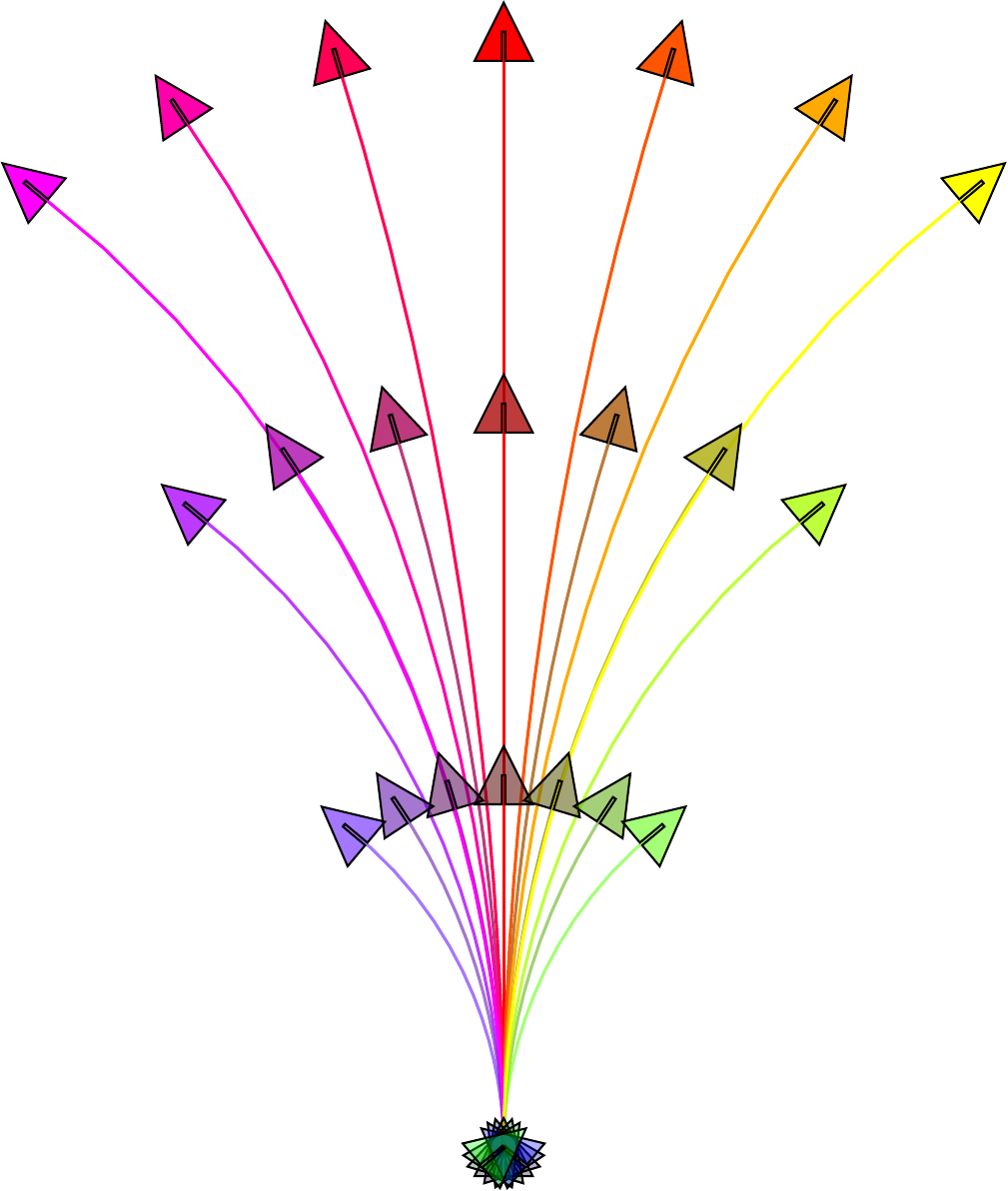}
\caption{\label{fig:actionspace}\textbf{The action space}: 28 actions, 4 choices of linear velocities $\in[0,1]$ m/s, 7 choices for angular vel. $\in[-3,3]$ rad/s. Arrows show the effect on pose of action held for $\frac{2}{3}$sec.}
\end{wrapfigure}
\myparagraph{Adding realism} 
To be able to smooth robot trajectories, we need to model accelerations. We extend the agent state as:
\begin{equation}
\mathbf{x}_t = [ x_t, y_t, \theta_t, v_t, w_t, \dot{v}_t, \dot{w}_t ],    
\end{equation}
where $x_t, y_t, \theta_t$ are the absolute position and orientation of the robot in the plane of the scene (in m, m and rad);
$v_t, w_t$ are the linear (forward) and angular velocities of the robot (in m/s and rad/s); 
$\dot{v}_t; \dot{w}_t$ are the linear and angular accelerations of the robot (in m/s$^2$ and rad/s$^2$).

The proposed agent predicts a pair $\langle v^*_t, w^*_t \rangle$ of velocity commands, 
chosen from a discrete action space $\mathcal{U}_\text{vel}$ resulting from the Cartesian product of linear and angular spaces {\small $\{0,0.3,0.6,1\}\times \{-3,-2,-1,0,\dots,3\}$}, as shown in Figure \ref{fig:actionspace}. To simplify notations, we will use $\mathbf{u}_t=\langle v^*_t, w^*_t\rangle\in\mathcal{U}$.

\begin{figure*}[t] \centering
\includegraphics[width=\linewidth]{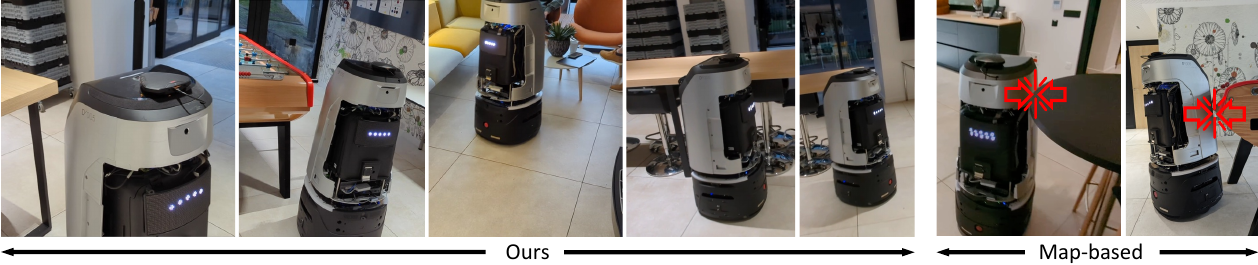}
\caption{\label{fig:obstacles}\textbf{Robustness} --- Left: the end-to-end trained agent is surprisingly robust and allows navigation very close to finely structured obstacles. Right: Navigation based on single ray Lidar and planning on occupancy maps, widely used in ROS based solutions, is difficult and error prone in situations where obstacles are thin at the height of the Lidar ray, are undetected and lead to collisions.}
\end{figure*}

One possibility to model realistic robot behavior is to design a full dynamical model of the robot in the form of a set of differential equations, identify its parameters, discretize it, and simulate it in Habitat, together with running an exact copy of the control algorithm used on the physical robot with exactly the same parameters and control frequency, which is a stringent constraint. Instead, we opted for a more flexible solution and we approximate the combined behavior of the physical robot and its control algorithm by a second order dynamical model~\cite{MondernControlEngineering2010}. We decompose interactions with the simulator into two different loops running at different frequencies:
\begin{itemize}[itemsep=1mm,labelindent=2mm,leftmargin=4mm,topsep=1mm]
    \item Visual observations, simulator state updates and agent decisions (Equations (\ref{eq:gru}) to (\ref{eq:supervision})) are produced in a slower loop, indexed by letter $t$ in the subsequent notation, and run at $f^*=3$Hz in our experiments.
    \item Between two subsequent steps of the slower loop, a faster loop models physical motion of the robot, without rendering observations. At the end of this loop, the simulator state is updated, and a visual observation is returned. Steps in this faster loop are indexed by a second index $\tau$ in the subsequent notation, $\mathbf{x}_{t,\tau}$.
\end{itemize}
In-between two environment steps $t{-}1$ and $t$, we simulate dynamics at frequency $f^\phi = 30 > f^*$. The number of physical sub-time steps per environment time step is therefore $K^\phi=\left\lceil\frac{f^\phi}{f^*}\right\rceil = 10$, their duration is $\Delta^\phi = \frac{1}{f^\phi} = 33$ms. 
The physical loop running between two environment steps can be formalized as the following set of state update equations,
\begin{equation}
\arraycolsep=1.4pt
\begin{array}{rll}
\mathbf{o}_t & = O(\mathbf{x}_{t-1}).
& \textbf{Observation} 
\\
\mathbf{x}_{t-1,0}&= \mathbf{x}_{t-1}
& \textbf{Init}
\\
\dot{v}_{t,\tau+1} &= \dot{v}_{t,\tau} + \Delta^\phi(                                {f^v_{t,\tau}}^2\delta^v_{t,\tau}
                                - 2\zeta^v_{t,\tau} f^v_{t,\tau}\dot{v}_{t,\tau}
                        ) 
& \textbf{Upd. acc.}                        
\\
                        \dot{w}_{t,\tau+1} &= \dot{w}_{t,\tau} + \Delta^\phi(                                {f^w_{t,\tau}}^2\delta^w_{t,\tau}
                                - 2\zeta^w_{t,\tau} f^w_{t,\tau}\dot{w}_{t,\tau}
                        )
\\
v_{t,\tau+1} &= v_{t,\tau} + \Delta^\phi v_{t,\tau+1} 
& \textbf{Upd. vel.}                        
\\
w_{t,\tau+1} &= w_{t,\tau} + \Delta^\phi w_{t,\tau+1}
\\                    
\theta_{t,\tau+1} &= \theta_{t,\tau} + \Delta^\phi w_{t,\tau+1} 
& \textbf{Upd. pose}                        
\\
x_{t,\tau+1} &= x_{t,\tau} + \Delta^\phi v_{t,\tau+1} \cos\theta_{t,\tau+1} 
\\
y_{t,\tau+1} &= y_{t,\tau} + \Delta^\phi v_{t,\tau+1} \sin\theta_{t,\tau+1}
\\                        
\mathbf{x}_{t} &= \mathbf{x}_{t-1,K}
& \textbf{Final state} 
\\
\end{array}
\end{equation}
In these equations,  $f^.$
are natural frequencies and damping coefficients
of asymmetric 2\textsuperscript{nd} order dynamic models for linear and angular motion. At each step $\tau$, they are chosen from identified values given command errors as follows:
\begin{equation}
\arraycolsep=1.4pt
\begin{array}{rl}
\delta^v_{t,\tau} &= v^*_t - v_{t,\tau} 
\\
\langle f^v_{t,\tau}, \zeta^v_{t,\tau} \rangle &= \begin{cases}
        \langle f^{v\uparrow}, \zeta^{v\uparrow} \rangle &\text{if } \delta^v_{t,\tau}\cdot v_{t,\tau} > 0 \text{ (acceleration)} \\
        \langle f^{v\downarrow}, \zeta^{v\downarrow} \rangle &\text{otherwise (deceleration)}
\end{cases} \\
\end{array}
\end{equation}
and similarly for angular model parameters $\langle f^w, \zeta^w\rangle$.
Velocity and acceleration are also clipped to identified maximum values. 

\myparagraph{System identification} the model has 8 parameters, 
$f^{v\uparrow}, \zeta^{v\uparrow}, 
f^{v\downarrow}, \zeta^{v\downarrow}, 
f^{w\uparrow}$, $\zeta^{w\uparrow},
f^{w\downarrow}$, $\zeta^{w\downarrow}$, which we identify from trajectories of a real robot controlled with pre-computed trajectories exploring the command space, see the appendix for more details.

\myparagraph{Training} the model is trained with RL, in particular PPO~\cite{schulman2017proximal}, and with a reward inspired by \citep{chattopadhyay2021robustnav},
$
r_t=\mathrm{R} \cdot \mathbb{I}_{\text {success}} -\Delta_t^{\mathrm{Geo}} -\lambda -\mathrm{C} \cdot \mathbb{I}_\text{collision}
$, where $R{=}2.5$, $\Delta_t^{\mathrm{Geo}}$ is the gain in geodesic distance to the goal, a slack cost $\lambda{=}0.01$ encourages efficiency, and a cost $C=0.1$ penalizes each collision without ending the episode.

\myparagraph{Recovery behavior} similar to what is done in classical analytical planning, we added a rudimentary 
recovery behavior on top of the trained agent: if the agent is notified of an obstacle and does not move for five seconds, it will move backward at 0.2 m/s for two seconds.


\section{Experimental Results}
\label{sec:experiments}

\myparagraph{Experimental setup} we trained the agent in the Habitat simulator~\cite{Savva_2019_ICCV} on the 800 scenes of the HM3D  Dataset~\cite{ramakrishnan2021hm3d} for 200M env. steps. Real robot experiments have been performed on a \textit{Naver Rookie} robot, which came equipped with various sensors. We added an additional front facing RGB camera and capture images with a resolution of $1280{\times}720$ resized to $256{\times}144$. In our experiments, the Lidar scan $\mathbf{S}$ is not taken from an onboard Lidar, but simulated from the 4 \textit{RealSense} depth cameras which are installed close to the floor and oriented in 4 different directions. We use multiple scan lines of the cameras and fuse them into a single ray, details are given in the appendix. We did, however, add a single plane Lidar to the robot (which did not come equipped with one) and used it for localization only (see section \ref{sec:method}). All processing has been done onboard, thanks to a \textit{Nvidia Jetson AGX Orin} we added, with an integrated \textit{Nvidia Ampere} GPU. It handles pre-processing of visual observations and network forward pass in around 70ms. The exact network architecture of the policy is given in the appendix.

\myparagraph{Evaluation} evaluating sim2real transfer is inherently difficult, as it would optimally require to evaluate all agent variants and ablations on a real physical robot and on a high number of episodes. We opted for a three-way strategy, all tables are color-coded with numbers corresponding to one of the three following settings:
\tcbox[on line,colframe=white,boxsep=0pt,left=1pt,right=1pt,top=1pt,bottom=1pt,colback=green!20]{\textbf{(i) ``Real''}} experiments evaluate the agent trained in simulation on the real Naver Rookie robot. It is the only form of evaluation which correctly estimates navigation performance in a real world scenario, but for practical reasons we limit it to a restricted number of 20 episodes in a large office environment shown in Fig. \ref{fig:obstacles}.
\tcbox[on line,colframe=white,boxsep=0pt,left=1pt,right=1pt,top=1pt,bottom=1pt,colback=orange!20]{\textbf{(ii) ``Simulation (+dyn. model)'')}} is a setting in simulation (Habitat), which allows large-scale evaluation on a large number of unseen environments and episodes, the HM3D validation set. Most importantly, during evaluation the simulator uses the identified dynamical model and therefore realistic motion, even for baselines which do not have access to one during training. Similar to the ``Real'' setting, this allows to evaluate the impact of not modeling realistic motion during training.
\tcbox[on line,colframe=white,boxsep=0pt,left=1pt,right=1pt,top=1pt,bottom=1pt,colback=blue!20]{\textbf{(iii) `` Simulation (train domain)''}} evaluates in simulation with the same action space an agent variant uses during training, i.e. there is no sim2real gap at all. This setting evaluates the difficulty of the simulated task, which might be a severe approximation of the task in real conditions. High performance in this setting is not necessarily indicative of high performance in a real environment.

\begin{figure*}[tb] \centering
    \includegraphics[width=\linewidth]{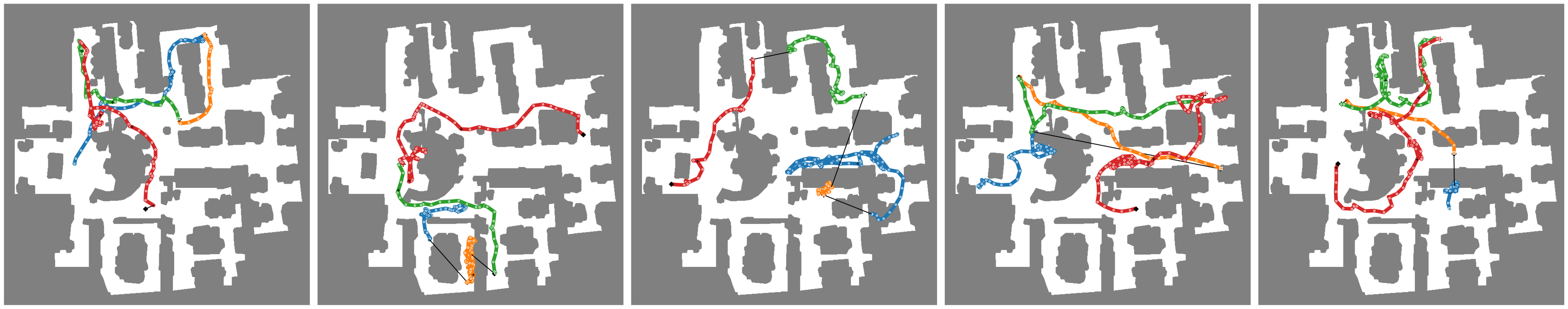}
    \caption{\label{fig:traj}\textbf{The 20 trajectories} of \epbox{Office/20} are distributed over 5 plots and superimposed on the map, color coded. If an episode fails, the remaining distance is shown as a straight black line from the last position to the goal. The agent is variant $\beta$ of Table~\ref{tab:ros}.
    \vspace*{-3mm}
    }
\end{figure*}

\begin{table*}[t] \centering
{\small 
\begin{tabular}{LLCCNNNNNNRRR}
\toprule
\rowcolor{TableGray2}
& \textbf{Method} & \textbf{+dyn} & \textbf{Ref.} &
\multicolumn{3}{c}{\cellcolor{orange!60}Sim (+dyn.) \epbox{HM3D/2.5k}} &
\multicolumn{3}{c}{\cellcolor{orange!60}Sim (+dyn.) \epbox{Office/20}} &
\multicolumn{3}{c}{\cellcolor{green!50}Real \epbox{Office/20}} 
\\
\rowcolor{TableGray2}
& \textbf{\& action space} & \textbf{(train)}&&
{\cellcolor{orange!60}\bf  SR(\%)} & {\cellcolor{orange!60}\bf  SPL(\%)} & {\cellcolor{orange!60}\bf SCT(\%)} &
{\cellcolor{orange!60}\bf  SR(\%)} & {\cellcolor{orange!60}\bf  SPL(\%)} & {\cellcolor{orange!60}\bf SCT(\%)} &
{\cellcolor{green!50}\bf  SR(\%)} & {\cellcolor{green!50}\bf  SPL(\%)} & {\cellcolor{green!50}\bf SCT(\%)} 
\\
\midrule
\textbf{(a)} & \textbf{Position, disc(4)} & \xmark & \cite{SadekICRA2022} &
{ 58.2} & { 48.2} & { 16.8} &
{ 35.0} & { 30.3} & { 13.2} &
15.0 & 11.8 & 2.4 
\\
\textbf{(b)} & \textbf{Velocity, disc(28)} & \xmark & \cite{yokoyama2021success} &
{ 0.8} & { 0.1} & { 0.1} &
{ 0.0} & { 0.0} & { 0.0} &
{ 20.0} & { 8.7} & { 2.6} 
\\
\textbf{(c)} & \textbf{Velocity, disc(28)} & \cmark & \textbf{(Ours)}&
{\bf 97.4} & {\bf 82.2} & {\bf 51.0} &
{\bf 100.0} & {\bf 78.8} & {\bf 59.9} &
{\bf 40.0} & {\bf 28.9} & {\bf 10.1}  
\\
\bottomrule
\end{tabular}
}
\caption{\label{tab:variants}\textbf{Impact of training with a realistic dynamical model:} we compare end-to-end trained models using position commands (a) as in \cite{SadekICRA2022} and, discretized velocity commands without a dynamical model (considering constant velocity) as in \cite{yokoyama2021success}, and our proposed method (c). The references \cite{SadekICRA2022,yokoyama2021success} are cited for their action space and motion handling, but they have different agent architectures.}
\vspace*{-1mm}
\end{table*}

We evaluate on two different sets of episodes:
\epbox{HM3D/2.5k} consists of 2500 episodes in the HM3D validation scenes, used in simulation only.
\epbox{Office/20} consists of 20 episodes in the targeted office building, Figure~\ref{fig:traj}. They are used for evaluation in both, real world and simulation.

\myparagraph{Metrics}
Navigation performance is evaluated by success rate (SR), i.e., fraction of episodes terminated within a distance of ${<}0.2$m to the goal by the agent calling the 
$\langle v{=}0,w{=}0\rangle$
action enough time to cancel its velocity, and SPL~\citep{DBLP:journals/corr/abs-1807-06757}, i.e., SR weighted by the optimality of the path,
$
\textit{SPL}=\frac{1}{N} \sum_{i=1}^N \mathbb{I}_{\text {success}} \frac{\ell_i^*}{\max (\ell_i, \ell_i^*)} ,
\label{eq:spl}
$
where $\ell_i$ is the agent path length and $\ell_i^*$ the GT path length.

SPL is limited in its ability to properly evaluate agents with complex dynamics. \textit{Success Weighted by Completion Time} (SCT)~\cite{yokoyama2021success} explicitly takes the agent's dynamics model into consideration, and aims to accurately capture how well the agent has approximated the fastest navigation behavior. It is defined as
$
\textit{SCT}=\frac{1}{N} \sum_{i=1}^N S_i \frac{t_i^*}{\max (c_i, t_i^*)},
\label{eq:sct}
$
where $c_i$ is the agent’s completion time in episode $i$, and $t^*_i$ is the shortest possible amount of time it takes to reach the goal point from the start point while circumventing obstacles based on the agent’s dynamics. To simplify implementation, we use a lower bound on $t^*$ taking into account linear dynamics along straight shortest path segments.

Checkpoints have been chosen as follows: performance in \tcbox[on line,colframe=white,boxsep=0pt,left=1pt,right=1pt,top=1pt,bottom=1pt,colback=green!20]{\textbf{Real}} is given with checkpoints selected as the ones providing max SR in \tcbox[on line,colframe=white,boxsep=0pt,left=1pt,right=1pt,top=1pt,bottom=1pt,colback=orange!20]{\textbf{simulation}}. Performance in \tcbox[on line,colframe=white,boxsep=0pt,left=1pt,right=1pt,top=1pt,bottom=1pt,colback=orange!20]{\textbf{Simu}}\hspace{-1mm}\tcbox[on line,colframe=white,boxsep=0pt,left=1pt,right=1pt,top=1pt,bottom=1pt,colback=blue!20]{\textbf{lation}} is given on the last checkpoint.

\begin{table}[t] \centering
{\small 
\setlength{\tabcolsep}{1pt}
\begin{tabular}{CLSSSNNN}
\toprule
\rowcolor{TableGray2}
& \textbf{Method} &
\multicolumn{3}{c}{\cellcolor{blue!50}Sim(train)\epbox{HM3D/2.5k}} &
\multicolumn{3}{c}{\cellcolor{orange!60}Sim(+dyn)\epbox{HM3D/2.5k}} 
\\
\rowcolor{TableGray2}
&\textbf{}&
{\cellcolor{blue!50}\bf  SR\%} & {\cellcolor{blue!50}\bf  SPL\%} & {\cellcolor{blue!50}\bf SCT\%} &
{\cellcolor{orange!60}\bf  SR\%} & {\cellcolor{orange!60}\bf  SPL\%} & {\cellcolor{orange!60}\bf SCT\%} 
\\
\midrule
\textbf{(a)} & \textbf{Pos, disc(4)} &
{ 92.7} & { 81.7} & { 30.3} &
{ 58.2} & { 48.2}	& { 16.8}	 
\\
\textbf{(b)} & \textbf{Vel, disc(28)} &
{\bf 98.0} & { 74.2} & {\bf 58.9} &
{  0.8} & {  0.1} & {  0.1}	
\\
\textbf{(c)} & \textbf{Vel, disc(28)} &
{ 97.4} & {\bf 82.2} & { 51.0} &
{\bf 97.4} & {\bf 82.2} & {\bf 51.0}
\\
\bottomrule
\end{tabular}
}
\caption{\label{tab:difficulty}\textbf{Difficulty of tasks given action spaces and motion models}: we evaluate the baseline model variants in the same simulated environment in which they have been trained. High performance does not necessarily transfer to real. Letters are variants in Table \ref{tab:variants}.}
\vspace*{-1mm}
\end{table}

\subsection{Results}

\myparagraph{Agent behavior} the agent is surprisingly robust and does not collide with the infrastructure even though it can quite closely approach it navigating around it, even if the obstacles are thin and light. Examples are given in Figure~\ref{fig:obstacles} (left). This is in stark contrast to the ROS 2 based planner, which uses a 2D Map constructed with the onboard Lidar: thin and light structures do not show up on the map or a filtered, often because the larger part of the obstacle is not in the height of the single Lidar ray. Collisions are frequent, examples are given in Figure \ref{fig:obstacles} (right).

\myparagraph{Sim2real gap and memory} one interesting finding we would like to share is that the raw base agent sometimes tends to get discouraged from initially being blocked in a situation, for instance if the passage through a door is not optimal and requires correction. The initial variants of the agent seemed to abandon relatively quickly and started searching for a different trajectory, circumventing the passage way obstacle altogether. We conjecture that this stems from the fact that such blockings are rarely seen in simulation and the agent is trained to ``write off'' this area quickly, storing in its recurrent memory that this path is blocked. All our real experiments are therefore performed with a version, which resets its recurrent state $\mathbf{h}_i$ (eq. (\ref{eq:gru})) every 10 seconds, leading to less frequent abandoning. Future work will address this problem more directly, for instance by simulating blocking situations during training.

\myparagraph{Influence of motion model}
Table \ref{tab:variants} compares results of different agents trained with different action spaces and with or without realistic motion during simulation. We can see that the impact of training the agent with the correct motion model in simulation is tremendous. The agent in line (b) uses the same action space, but no dynamical model is used in simulation, which means that changes in velocity are instantaneous and velocities are constant between decisions. The behavior is unexploitable, the agent is disoriented and keeps crashing into its environment. Line (a) corresponds to a position controlled agent with action space $\{${\small \texttt{FORWARD 25cm}, \texttt{TURN\_LEFT $10^{\circ}$}, \texttt{TURN\_RIGHT $10^{\circ}$}, \texttt{STOP}}$\}$. After training, it is adapted to motion commands by calculating the corresponding velocity commands given the decision frequency of 3 Hz. It has somewhat acceptable (but low) performance in simulation, although it does not dispose of a motion model during training. This fact that rotating and linear motion are separated and cannot occur in the same time simplifies the problem and leads to some success in simulation, but this behavior does not transfer well to the real robot. The agent with the identified motion model achieves near perfect SR in simulation, which shows that the model can learn to anticipate motion and (internal latent) future state prediction correctly. When transferred to the real robot, performance is the best among all agents, but still leaves room much room for improvement.

\myparagraph{Difficulty of the tasks}
Table \ref{tab:difficulty} compares the same three agents also in simulation when validated with the same setting they have been trained on (action space, motion model or absence of). While this comparison is not at all indicative of the performance of these agents on a real robot, it provides evidence of the difficulty of the task in simulation. Surprisingly, the performances are very close: the additional burden the motion model puts on the learning problem itself can be handled very well by the agent.

\begin{table}[t] \centering
{\small 
\setlength{\tabcolsep}{1pt}
\begin{tabular}{LCCNNNRRR}
\toprule
\rowcolor{TableGray2}
\textbf{Point Goal} & 
$\hat{\mathbf{p}}^r_t$ & 
$\hat{\mathbf{p}}^a_t$ &
\multicolumn{3}{c}{\cellcolor{orange!60}Sim (+dyn)\epbox{Office/20}} &
\multicolumn{3}{c}{\cellcolor{green!50}Real \epbox{Office/20}} 
\\
\rowcolor{TableGray2}
&&&
{\cellcolor{orange!60}\bf  SR\%} & {\cellcolor{orange!60}\bf  SPL\%} & {\cellcolor{orange!60}\bf SCT\%} &
{\cellcolor{green!50}\bf  SR\%} & {\cellcolor{green!50}\bf  SPL\%} & {\cellcolor{green!50}\bf SCT\%} 
\\
\midrule
\textbf{$\mathbf{g}_t$} &\xmark&\xmark&
{ 40.0} & { 34.5} & { 27.4} &
35.0 & 23.6 & 5.2 
\\
\textbf{$\mathbf{g}_0$} &\cmark&\cmark& 
{ 70.0} & { 51.9} & { 36.9} &
{\bf 50.0} & {\bf 37.5} & {\bf 11.4}  
\\
\textbf{$\mathbf{g}_0$+ superv.} &\cmark&\cmark& 
{\bf 100.0} & {\bf 78.8} & {\bf 59.9} &
40.0 & 28.9 & 10.1
\\
\textbf{$\mathbf{g}_0$+ superv.} &\cmark&\xmark& 
70.0 & 51.9 & 36.9 &
25.0 & 16.7 &  3.0
\\
\bottomrule
\end{tabular}
}
\caption{\label{tab:pointgoal}\textbf{Localization and PointGoal calculation}: we compare the impact of the presence of external localization to the agent, and the point goal sources: dynamical point goal (through external localization) with static point goal w/ and w/o supervision. All agents are variant (c) from Table \ref{tab:variants}.}
\vspace*{-1mm}
\end{table}

\myparagraph{Goal vector calculation} 
As explained in Section \ref{sec:method}, our base agent receives the static point goal $\mathbf{g}_0$, ie. a vector with respect to the initial reference frame at time $t{=}0$, which is not updated. Table \ref{tab:pointgoal} compares this agent with two other variants. A version where supervision is removed during training, which surprisingly shows good performance. We conjecture that the additional supervision learns integration of odometry which might not transfer well enough from simulation, indicating insufficient noise modeling in simulation. This will be addressed in future work. In another variant the dynamic point goal $\mathbf{g}_t$ is provided at each time step in the agent's egocentric frame. It is calculated from an external localization source, noisy in simulation. Letting the agent itself take care of the point goal integration seems to be the better solution.

\begin{table}[t] \centering
{\small 
\setlength{\tabcolsep}{1pt}
\begin{tabular}{LRRR|RRR}
\toprule
\rowcolor{TableGray2}
\textbf{Method} & 
\multicolumn{3}{c|}{\cellcolor{green!50}Real \epbox{Office/20}} &
\multicolumn{3}{c}{\cellcolor{green!50}Real \epbox{Office/20-alt}} 
\\
\rowcolor{TableGray2}
&
{\cellcolor{green!50}\bf  SR\%} & {\cellcolor{green!50}\bf  SPL\%} & {\cellcolor{green!50}\bf SCT\%} &
{\cellcolor{green!50}\bf  SR\%} & {\cellcolor{green!50}\bf  SPL\%} & {\cellcolor{green!50}\bf SCT\%} 
\\
\midrule
($\alpha$) \textbf{ROS 2 (fused)} & 
\textbf{ 80.0} & {\bf 69.5} & {\bf 26.2} &
\textbf{ 90.0} & {\bf 70.5} & {\bf 27.0} 
\\
($\beta$) \textbf{Ours (finetuned)} & 
{ 55.0} & { 40.4} & { 11.2} &
{ 60.0} & { 42.0} & { 9.7} 
\\
\bottomrule
\end{tabular}
}
\caption{\label{tab:ros}\textbf{Comparisons} --- we compare with the ROS 2 NavStack, which has access to the 2D occupancy map beforehand, localizes itself with the single ray Lidar scan and AMCL (Monte Carlo localization), uses the fused laserscan for obstacle avoidance, followed by shortest path planning. \textbf{Metrics do no measure collisions}, which are much higher for the ROS 2 planner. To be comparable, our method is finetuned with RL on the Matterport scan of the same building. We also add an experiment on the same office building with a different furniture arrangement, \epbox{Office/20-alt}.
}
\vspace*{-1mm}
\end{table}

\myparagraph{Comparison with a map based planner}
Table \ref{tab:ros} compares the method with a ROS 2 based planner, which uses a 2D occupancy map constructed beforehand (not on the fly). To be comparable, we finetuned our agent on a Matterport scan of the same building. While the ROS 2 based planner is still more efficient in terms of SR and time (SCT), it requires many experiments to finetune its parameters. For example, low values for the inflation radius will produce many collisions with tables. But when this parameter is too high, the planner cannot find any path through doors and corridors.

\myparagraph{Rearrangement} we test the impact of rearranging furniture significantly in the Office environment (see the appendix for pictures) and show the effect in Table \ref{tab:ros}, block \epbox{Office/20-alt}. The differences are neglectable.

\begin{figure} \centering
    \includegraphics[width=\linewidth]{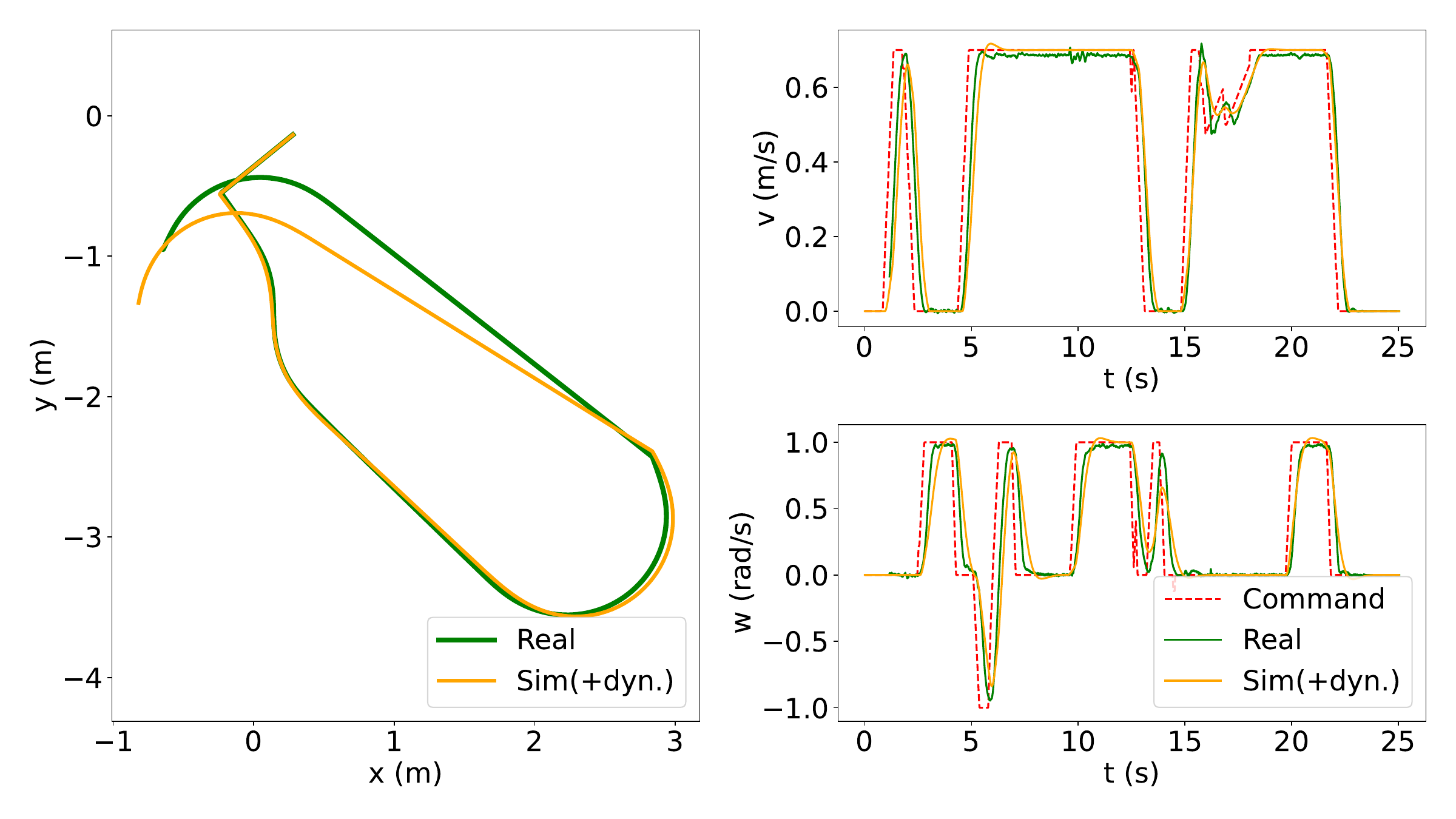}
    \caption{\label{fig:motionmodel}\textbf{Dynamical sim2real gap:} we compared recorded trajectories to simulated rollouts obtained with the same actions.}
    \vspace*{-3mm}
\end{figure}

\myparagraph{Dynamical sim2real gap}
Figure \ref{fig:motionmodel} compares real robot trajectories obtained by the agent on a small map of $4m{\times}4$m with rollouts of the motion model in simulation with the same actions, indicating very small drift. The dynamical models seems to approximate real robot motion very well.

\begin{figure}[t] \centering
\includegraphics[width=\linewidth]{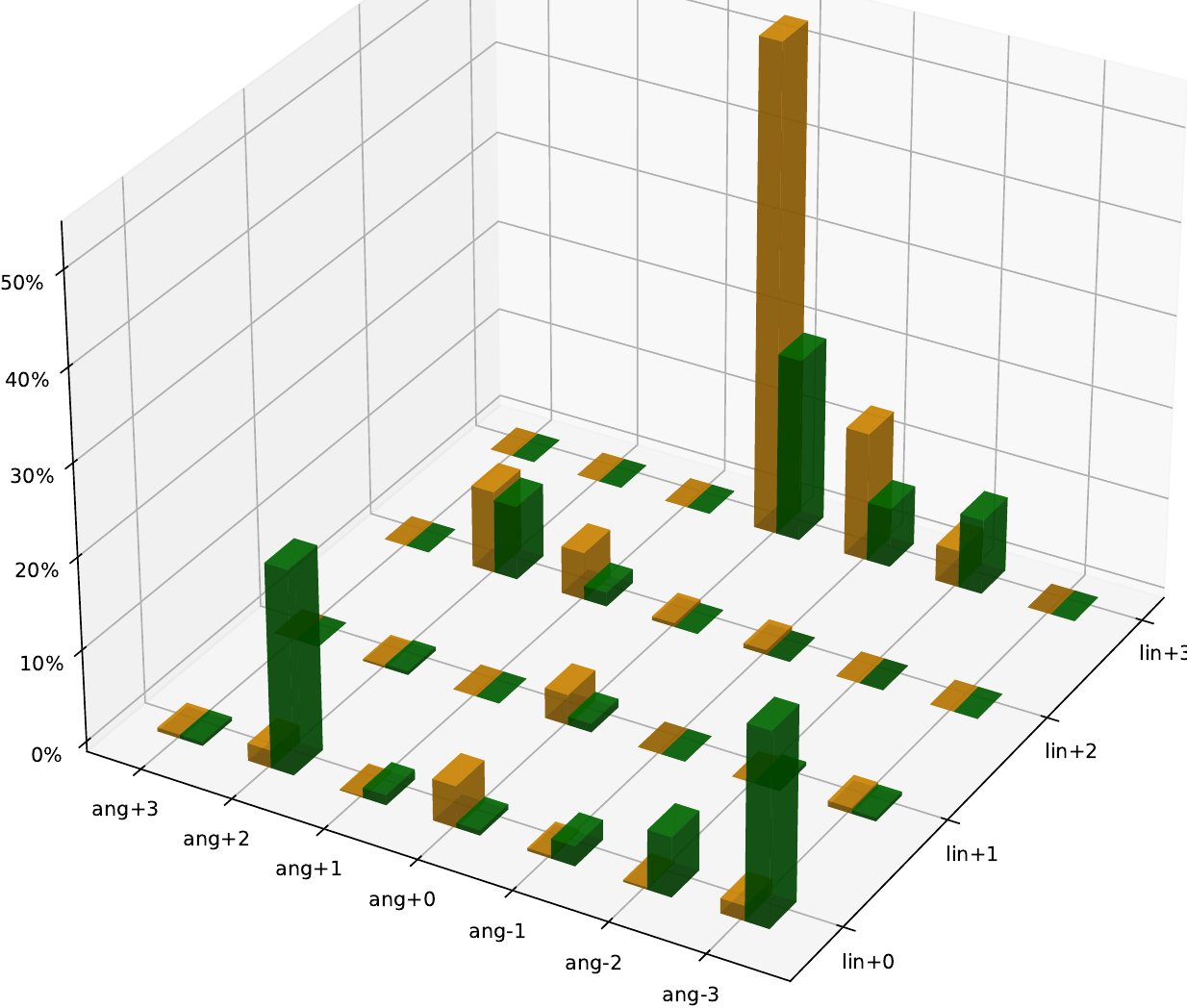}
\caption{\label{fig:actiondistribution} \textbf{Distribution of actions}, \epbox{Office/20},{\color{ColBarSim}\rule[0.8mm]{4mm}{0.7mm}} Sim, {\color{ColBarReal}\rule[0.8mm]{4mm}{0.7mm}} Real.}
\end{figure}
\myparagraph{Actions taken}
Figure \ref{fig:actiondistribution} shows histograms of the actions taken by the action in simulation and in real on \epbox{Office/20}. There is a clear preference for certain combinations of linear and angular velocity. Distributions in sim and real mostly match, except for a tendency to turn in-place in real, which could be explained by obstacle mis-detections due to scan range noise.

\section{Conclusion}

\noindent
We have presented an end-to-end trained method for swift and precise navigation which can robustly avoid even thin and finely structured obstacles. This is achieved with training in simulation only by adding a realistic motion model identified from recorded trajectories from a real robot. The method has been extensively evaluated in simulation as well as on a real robotic platform, where we assessed the impact of the motion model, the action space, and the way how a point goal is calculated and provided to the agent. The method is robust, future work will focus on closed-loop adaptation of dynamics and sensor noise estimation.

{
    \small
    \bibliographystyle{ieeenat_fullname}
    \bibliography{ms,zotero_notlinked,ms_new_cvpr2024}
}

\appendix

\vspace*{4mm}

\noindent
{\huge \textbf{Appendix}}

\section{System identification}
\label{sec:system-id}

We identified the parameters of the dynamical model from $N{=}8$ recorded trajectories of various lengths $K_n$ on the real robot
using \texttt{ros2 bag} while executing different command steps patterns:
$$\Big\{t_{n,k},v^*_{n,k},w^*_{n,k},v_{n,k},w_{n,k}\ \forall k\in\{1..K_n\}\Big\}_{n\in\{1..N\}},$$
where $t_{n,k}$ are time-stamps, $v^*_{n,k},w^*_{n,k}$ are velocity commands and $v_{n,k},w_{n,k}$ are odometry measurements
sampled at a frequency of $\frac{1}{t_{n,k+1} - t_{n,k}}\approx 100$Hz (commands are interpolated using zero-order hold to match odometry timestamps).

We smooth odometry signals using a \emph{Hanning} window of size $21$ ($\sim 210$ms),
then compute the first and second order derivatives
$\dot{v}_{n,k}, \ddot{v}_{n,k}$ (resp. $\dot{w}, \ddot{w}$)
by central finite difference:
$$\begin{array}{rlr}
	\dot{v}_{n,k} =& \frac{v_{n,k+1} - v_{n,k-1}}{t_{n,k+1} - t_{n,k-1}}, &\forall n,\forall k\in[2,K_n-1], \\
	\ddot{v}_{n,k} =& \frac{\dot{v}_{n,k+1} - \dot{v}_{n,k-1}}{t_{n,k+1} - t_{n,k-1}}, &\forall n,\forall k\in[3,K_n-2].
\end{array}$$

We compute the error $\delta^v_{n,k} = v^*_{n,k} - v_{n,k}$
and split the samples w.r.t. the sign of term $\delta^v_{n,k}  v_{n,k}$ in order to identify the acceleration ($\delta^vv > 0$) and deceleration ($\delta^vv < 0$) phases of the trajectory,
corresponding to parameters $f^{v\uparrow}, \zeta^{v\uparrow}$
(resp. $f^{v\downarrow}, \zeta^{v\downarrow}$) of our asymmetric second order models presented in Section 4 of the main paper.

We then solve $4$ different least square problems (\{linear,angular\}$\times$\{accel,decel\}), such as:
$$
	\begin{bmatrix}
		\vdots & \vdots \\
		\delta^v_{n,k} & \dot{v}_{n,k} \\
		\vdots & \vdots
	\end{bmatrix}
	\begin{bmatrix}
		{f^{v\uparrow}}^2 \\
		-2\zeta^{v\uparrow}f^{v\uparrow}
	\end{bmatrix}
	= \begin{bmatrix}
		\vdots \\
		\ddot{v}_{n,k} \\
		\vdots
	\end{bmatrix}_{\forall n,k : \delta^v_{n,k}v_{n,k} > 0}
$$
from which we can easily deduce best-fit values of $f^{v\uparrow}$ and $\zeta^{v\uparrow}$
(resp. $f^{v\downarrow}$, $\zeta^{v\downarrow}$, $f^{w\uparrow}$, $\zeta^{w\uparrow}$, $f^{w\downarrow}$, $\zeta^{w\downarrow}$).
We also use these samples to define our saturation values:
$$\begin{array}{rl}
	v_\text{max} =& \max\{v_{n,k},\ \forall n, \forall k\} \\
	v_\text{min} =& \min\{v_{n,k},\ \forall n,\forall k\} \\
	|\dot{v}|^\uparrow_\text{max} =& \max\{|\dot{v}_{n,k}|,\ \forall n, \forall k : \delta^v_{n,k}v_{n,k} > 0\} \\
	|\dot{v}|^\downarrow_\text{max} =& \max\{|\dot{v}_{n,k}|,\ \forall n, \forall k : \delta^v_{n,k}v_{n,k} < 0\}
\end{array}$$
(resp. $w_\text{max}, w_\text{min}, |\dot{w}|^\uparrow_\text{max}, |\dot{w}|^\downarrow_\text{max}$).

\myparagraph{Manual adjustment}
The PID included in the inner control loop of the robot
which converts velocity commands to wheel speed to motor current inputs
improves the response time of the entire system
in a way a second order model cannot properly capture.
Hence, the value obtained by the automatic identification can lead to some under-damped models,
with oscillations, to keep up with the fast rise times.
We manually tweaked these values to improve damping and reduce oscillations,
setting all $\zeta=0.7$, and compensating for slower rise times
by increasing natural frequencies $f$ accordingly.

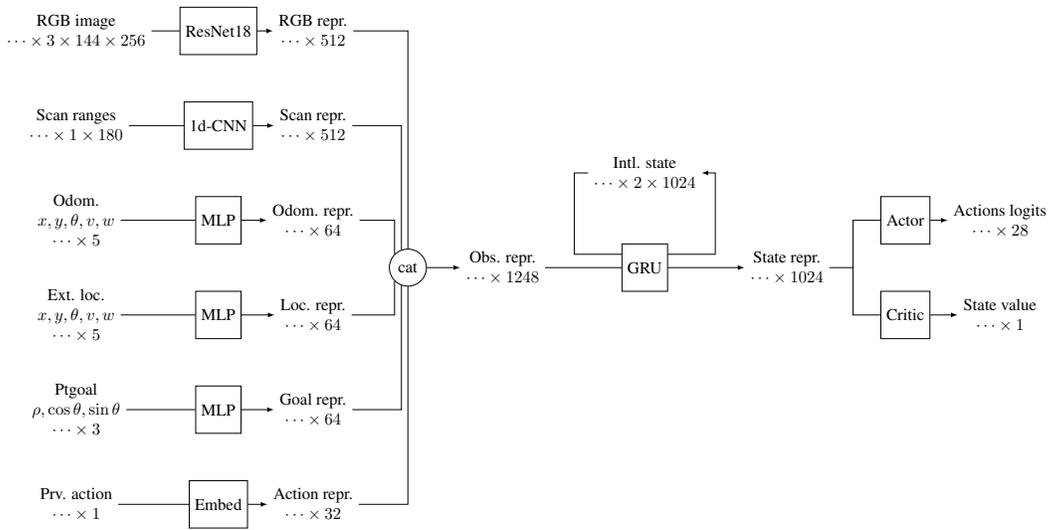
\begin{figure*}[ht]
\centering
\resizebox{0.8\linewidth}{!}{\begin{tikzpicture}[every node/.style={align=center}, nnmod/.style={draw,minimum height=1cm}]
	\node (rgb) at (0, 0) {RGB image \\ $\dots\times 3\times 144\times 256$};
	\node (scan) at (0, -2) {Scan ranges \\ $\dots\times 1\times 180$};
	\node (odom) at (0, -4) {Odom. \\ $x,y,\theta,v,w$ \\ $\dots\times 5$};
	\node (loc) at (0, -6) {Ext. loc. \\ $x,y,\theta,v,w$ \\ $\dots\times 5$};
	\node (goal) at (0, -8) {Ptgoal \\ $\rho,\cos\theta,\sin\theta$ \\ $\dots\times 3$};
	\node (act) at (0, -10) {Prv. action \\ $\dots\times 1$};

	\node[nnmod] (rn18) at (3, 0) {ResNet18};
	\draw (rgb) -- (rn18);
	\node[nnmod] (cnn1d) at (3, -2) {1d-CNN};
	\draw (scan) -- (cnn1d);
	\node[nnmod] (odom mlp) at (3, -4) {MLP};
	\draw (odom) -- (odom mlp);
	\node[nnmod] (loc mlp) at (3, -6) {MLP};
	\draw (loc) -- (loc mlp);
	\node[nnmod] (goal mlp) at (3, -8) {MLP};
	\draw (goal) -- (goal mlp);
	\node[nnmod] (emb) at (3, -10) {Embed};
	\draw (act) -- (emb);

	\node (rgb f) at (5, 0) {RGB repr. \\ $\dots\times 512$};
	\draw[-latex] (rn18) -- (rgb f);
	\node (scan f) at (5, -2) {Scan repr. \\ $\dots\times 512$};
	\draw[-latex] (cnn1d) -- (scan f);
	\node (odom f) at (5, -4) {Odom. repr. \\ $\dots\times 64$};
	\draw[-latex] (odom mlp) -- (odom f);
	\node (loc f) at (5, -6) {Loc. repr. \\ $\dots\times 64$};
	\draw[-latex] (loc mlp) -- (loc f);
	\node (goal f) at (5, -8) {Goal repr. \\ $\dots\times 64$};
	\draw[-latex] (goal mlp) -- (goal f);
	\node (act f) at (5, -10) {Action repr. \\ $\dots\times 32$};
	\draw[-latex] (emb) -- (act f);

	\node[draw, circle] (cat) at (7, -5) {cat};
	\draw (rgb f) -| (cat);
	\draw (scan f) -| (cat.110);
	\draw (odom f) -| (cat.135);
	\draw (loc f) -| (cat.-135);
	\draw (goal f) -| (cat.-110);
	\draw (act f) -| (cat);
	\node (obs f) at (9, -5) {Obs. repr. \\ $\dots\times 1248$};
	\draw[-latex] (cat) -- (obs f);

	\node[nnmod] (gru) at (12, -5) {GRU};
	\draw (obs f) -- (gru);
	\node (intl) at (12, -3) {Intl. state \\ $\dots\times 2\times 1024$};
	\draw (intl.west) -| ($(gru.150)+(-1,0)$) -- (gru.150);
	\draw[-latex] (gru.30) -- ($(gru.30)+(1,0)$) |- (intl.east);
	\node (s f) at (15, -5) {State repr. \\ $\dots\times 1024$};
	\draw[-latex] (gru) -- (s f);

	\node[nnmod] (actor) at (17.5,-4) {Actor};
	\draw (s f) -- ($(s f.east)+(0.5,0)$) |- (actor);
	\node (logp) at (19.5,-4) {Actions logits \\ $\dots\times 28$};
	\draw[-latex] (actor) -- (logp);

	\node[nnmod] (critic) at (17.5,-6) {Critic};
	\draw ($(s f.east)+(0.5,0)$) |- (critic);
	\node (v) at (19.5,-6) {State value \\ $\dots\times 1$};
	\draw[-latex] (critic) -- (v);
\end{tikzpicture}}
\caption{Agent architecture overview}
\label{fig:arch}
\end{figure*}

\section{Agent architecture}

\myparagraph{An Overview} of the agent's network architecture is illustrated on Figure~\ref{fig:arch},
providing all intermediate representation sizes.
In total, the entire network has $25,589,541$ trainable parameters.

\myparagraph{The RGB encoder} is a standard ResNet18 with $64$ base planes,
and $4$ layers of $2$ basic blocks each.

\myparagraph{The scan encoder} is a simple 1d-CNN composed of $3$ \texttt{(Conv, ReLU, MaxPool)} blocks
with respectively $64$, $128$, and $256$ channels,
kernels of sizes $7$, $3$ and $3$,
circular paddings of sizes $3$, $1$ and $1$,
and pooling windows of widths $3$, $3$ and $5$,
followed by a Linear layer projecting the final flattened channels to a vector of size $512$.

\myparagraph{Odom, ext loc, and goal encoders}
are MLPs with a single hidden layer of size $1024$,
an output layer of size $64$,
and \texttt{ReLU} activations.

\myparagraph{The action encoder} is a discrete set of $29$ ($28$ actions plus one \texttt{NO\_ACTION} token)
embedding vectors of size $32$.

\myparagraph{The state encoder}
is a GRU module with $2$ layers, each having an internal recurrent state vector of size $1024$.

\begin{figure}[t]
\includegraphics[width=\columnwidth]{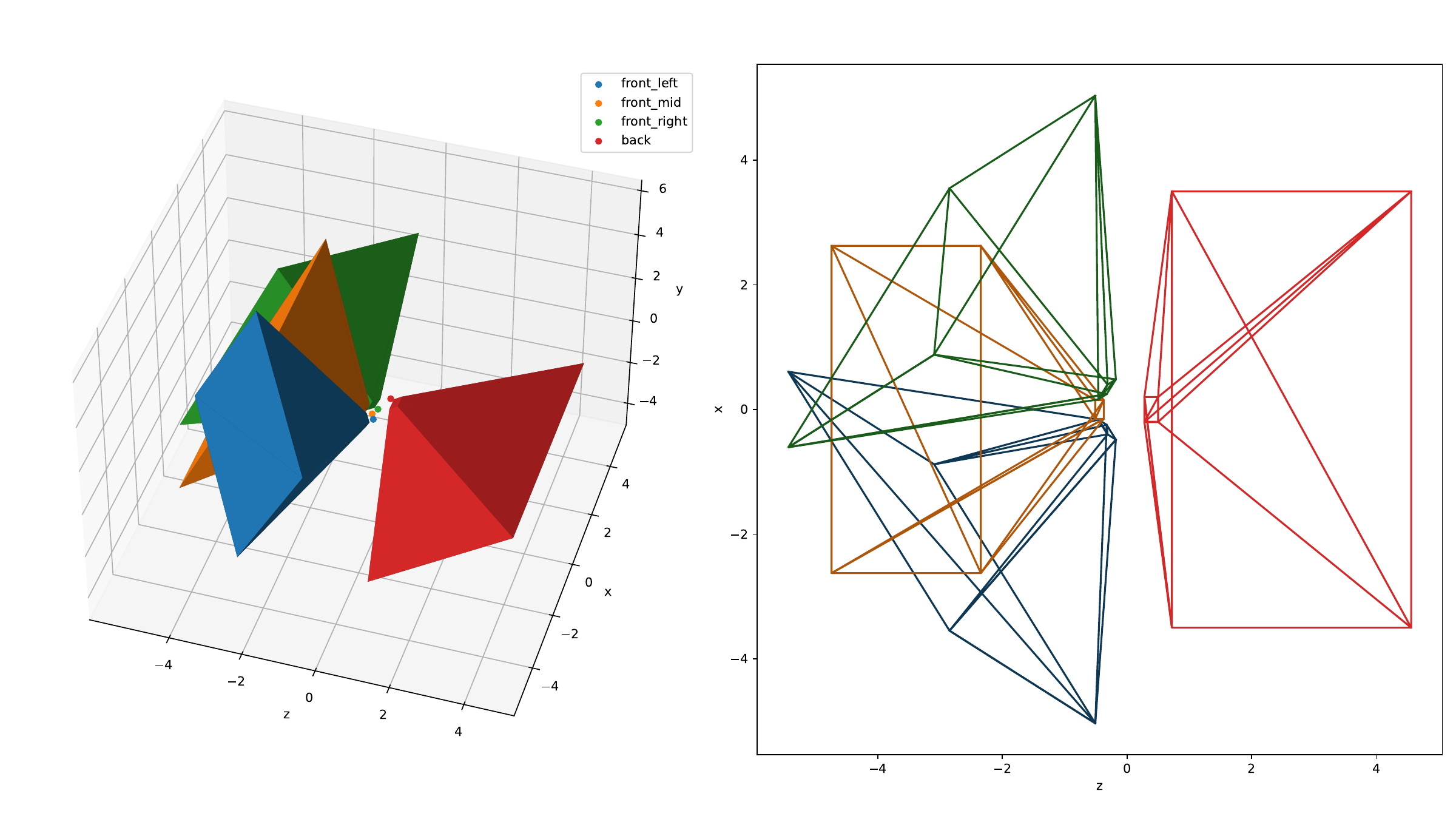}
\caption{Configuration of depth sensors to create fused scan.}
\label{fig:sensors}
\end{figure}

\section{Details on the fused simulated Lidar scan}

As mentioned in the main paper, we use the Lidar of the robot only for AMCL localization. The agent itself receives a scan which is similar to Lidar scan in its data representation, but is calculated from depth images. Four RealSense depth sensors are positioned around the robot as illustrated on Figure~\ref{fig:sensors}. In simulation they are rendered with 
with a resolution of $80\times 60$ pixels and a $90^\circ$ horizontal field of view. However, three sensors on the front are rotated $90^\circ$ along their optical axis
(i.e. in ``portrait-mode'') such that they have a larger vertical than horizontal field of view.

The (known) camera intrinsics and extrinsics are used to transform depth frames into pointclouds, where each pixel is mapped to local 3D world coordinates (relative to the robot's base). The pointclouds are then filtered in such a way that only points with height coordinates $y$ between $5$cm and $1.2$m are considered as obstacles to avoid (values outside this range are considered as navigable space where the robot can pass).

The $x$ and $z$ coordinates of cloudpoints are projected on the ground plane (ignoring height coordinate $y$), converted to polar coordinates and grouped by azimuth in $180$ bins, each bin of size $2^\circ$, covering the full range from $-180^\circ$ to $180^\circ$. The scan ranges are obtained by taking the minimum radius in each of the $180$ bins.

Optionally, a rolling buffer is used to maintain $N$ last pointclouds and to create one fused scan. 
In that case, exact robot motion is used to register the old pointclouds into the new robot reference frame.

\section{Furniture rearrangement}

\begin{figure}[t] \centering
\includegraphics[width=0.9\linewidth]{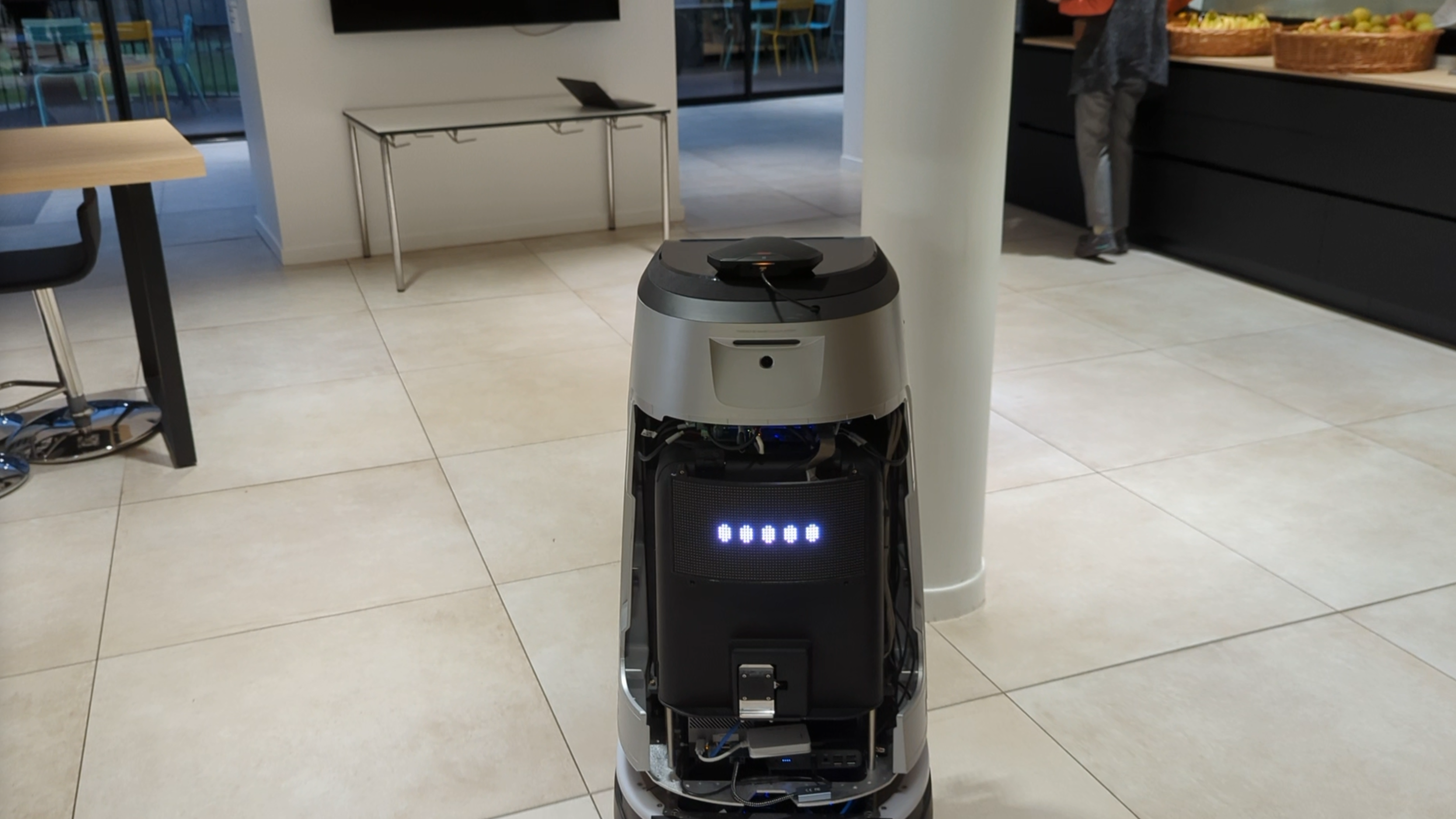}
\includegraphics[width=0.9\linewidth]{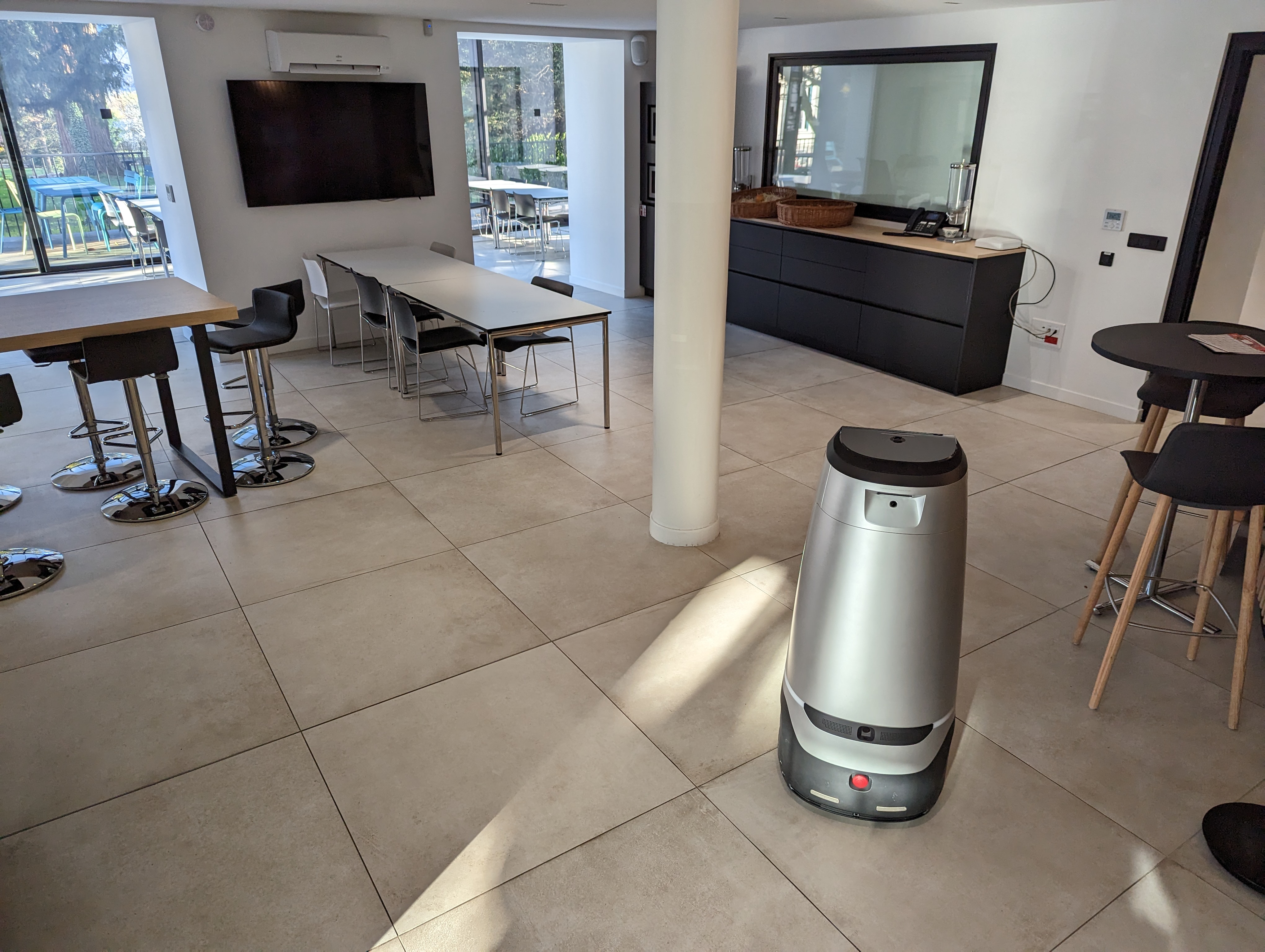}
\caption{\label{fig:arrangement} The scene configurations with two different tested furniture configurations. Top: \epbox{Office/20}. Bottom: \epbox{Office/20-alt}.}
\end{figure}

Figure~\ref{fig:arrangement} shows the differences between the arrangement of the office scene during the \epbox{Office/20} and the \epbox{Office/20-alt} experiments.

\end{document}